\documentclass{article}
\usepackage{spconf,amsmath,graphicx,hyperref}
\usepackage{amsmath}
\usepackage{amssymb}  
\usepackage{comment}
\usepackage{multirow}
\usepackage{graphicx}
\usepackage{subcaption}

\title{Adversarial Update-Based Federated Unlearning\\ for Poisoned Model Recovery}
\name{Wenwei Zhao$^{1}$, Xiaowen Li$^{2}$, Yao Liu$^{2}$, Zhuo Lu$^{1}$}
\address{$^{1}$Department of Electrical Engineering, University of South Florida,Tampa, FL, USA\\
$^{2}$Bellini College, University of South Florida, Tampa, FL, USA}
\begin{document}
%\ninept
%
\maketitle
\begin{abstract}
Federated learning (FL) is vulnerable to poisoning attacks, where malicious clients upload manipulated updates to degrade the performance of the global model. Although detection methods can identify and remove malicious clients, the model remains affected. Retraining from scratch is effective but costly, and existing unlearning methods remain unsatisfactory in both effectiveness and efficiency. We propose Federated Adversarial Unlearning (FAUN), a lightweight framework that retains only a short window of malicious clients' updates and employs adversarial optimization on a proxy dataset to derive updates that eliminate malicious directions. Applying these updates for a few unlearning rounds, followed by benign fine-tuning, enables fast removal of malicious effects and stable recovery. Experiments on three canonical datasets show that FAUN achieves recovery comparable to retraining while requiring far fewer rounds and reduces attack success rates to near zero, confirming FAUN successfully eliminates the contributions of unlearned clients.
\end{abstract}

\section{Introduction}
Federated learning (FL) has emerged as a promising paradigm that enables collaborative model training across distributed clients without exposing their raw data\cite{li2020federated}. However, the decentralized nature of FL also makes it highly vulnerable to poisoning attacks\cite{fang2020local, baruch2019little, cao2022mpaf}, where malicious clients deliberately manipulate their local updates to degrade the global model. To mitigate such threats, a lot of research has focused on detecting malicious clients and removing them from the FL system\cite{zhang2022fldetector, yan2023defl, krauss2023mesas}. However, even after detection, the global model has already been contaminated, and the server must recover an accurate model from the poisoned one.

The most straightforward solution is to retrain from scratch using only retained clients. While effective, this approach is computationally prohibitive, as it discards all previous progress and incurs substantial training cost. To address this inefficiency, recent works have explored the concept of federated unlearning, which aims to remove the contributions of detected malicious clients without requiring a restart of training\cite{cao2023fedrecover, liu2021federaser, jiang2024towards}.
Despite progress, existing unlearning methods still suffer from important limitations. Existing unlearning approaches, ranging from retraining to trajectory reconstruction and correction, remain unsatisfactory in federated settings. Retraining-based methods\cite{liu2021federaser, su2023asynchronous} are prohibitively expensive, while correction techniques\cite{cao2023fedrecover} and reconstruction\cite{liu2021revfrf} either rely on impractical complete historical storage or treat unlearning as a numerical rollback, often leaving residual malicious influence. These limitations call for a more efficient and principled solution.

In this work, we present Federated Adversarial Unlearning (FAUN), a novel method that efficiently recovers accurate global models from poisoning by using adversarially optimized updates for elimination. Instead of storing complete histories, we cache only the last few rounds of malicious clients' updates around detection and construct adversarial updates for elimination, thereby reducing storage costs and enabling fast recovery within several unlearning rounds.
We approximate the malicious contribution from each detected client's recent updates and amplify their worst-case poisoning effect via adversarial optimization on a small proxy dataset. These adversarial updates are then aggregated into an elimination direction and subtracted from the global model, thereby canceling malicious influence and steering the global model toward minimizing the loss induced by retained clients under a min-max principle of adversarial optimization\cite{wang2021adversarial}.
To avoid over-elimination, the adversarial elimination rounds are restricted to only a few rounds, after which the server continues with standard training on benign clients. This design enables the rapid removal of malicious influence while ensuring the stable recovery of model accuracy.

In summary, our contributions are threefold: (1) we propose Federated Adversarial Unlearning (FAUN), a framework that constructs adversarially optimized updates to counteract malicious contributions without costly retraining or trajectory replay; (2) we show that FAUN is lightweight, requiring only a short window of malicious clients' updates and a small proxy dataset, thus reducing storage and computation overhead; and (3) we demonstrate on multiple datasets with both untargeted and targeted poisoning that FAUN removes malicious effects within 2-10 elimination rounds, suppresses attack success rates to near zero, and achieves accuracy comparable to retraining with far fewer rounds and much lower overhead. These results highlight adversarial-update-based unlearning as an efficient solution for recovering federated learning from poisoning attacks.

\section{Preliminaries}
\subsection{Federated Learning (FL) and poisoning attack}
We consider a federated learning (FL) system with a client set $\mathcal{C} = \{1, \ldots, n\}$. At the $t$-th training round, the global model is denoted by $\mathbf{w}_t \in \mathbb{R}^d$. Each client $i \in \mathcal{C}$ trains on its local dataset $\mathcal{D}_i$ and computes the local gradient update $g_i^t = \nabla_{\mathbf{w}_t}\mathcal{L}(\mathbf{w}_t;\mathcal{D}_i)$, where $\mathcal{L}$ is the empirical loss function. The server aggregates the local gradients according to an aggregation rule $\mathcal{A}$\cite{fang2020local}. Then, the server updates the global model with a learning rate $\eta$, expressed as  
\begin{equation}
\mathbf{w}_{t+1} = \mathbf{w}_t - \eta \cdot \mathcal{A}(g_t^1, g_t^2, \ldots, g_t^n).
\end{equation}

During training, attacker can control a subset of clients and launch poisoning attacks by uploading manipulated local updates. These can be classified into untargeted poisoning attack\cite{fang2020local, shejwalkar2021manipulating, cao2022mpaf}, which reduces overall model accuracy, and targeted poisoning attack\cite{bagdasaryan2020backdoor, baruch2019little, bhagoji2019analyzing}, such as backdoor attacks that enforce specific misclassifications. In practice, the fraction of malicious clients is usually assumed to be less than half of the total population. Powerful detection techniques can identify malicious clients, after which training is stopped and the detected clients $\mathcal{C}_m \subseteq \mathcal{C}$ are removed from the FL system, leaving the benign clients $\mathcal{C}_r=\mathcal{C}\setminus\mathcal{C}_m$. Our goal is then to eliminate the contribution of $\mathcal{C}_m$ so that the global model can be restored without costly retraining from scratch.

\subsection{Server Requirements}
We assume that the server can store the global model and the local updates collected before malicious clients are detected. In addition, the server can manually collect a small clean proxy dataset, only used for the unlearning task. The server also has computational resources to leverage the proxy data for adversarial optimization and construct adversarial updates. These requirements are reasonable\cite{cao2020fltrust}, as the server is typically powerful in federated learning deployments.

\section{Methodology}
\begin{figure}[t]
  \centering
  \vspace{0mm} % Adjust the space above the figure
  \includegraphics[width=0.9\linewidth]{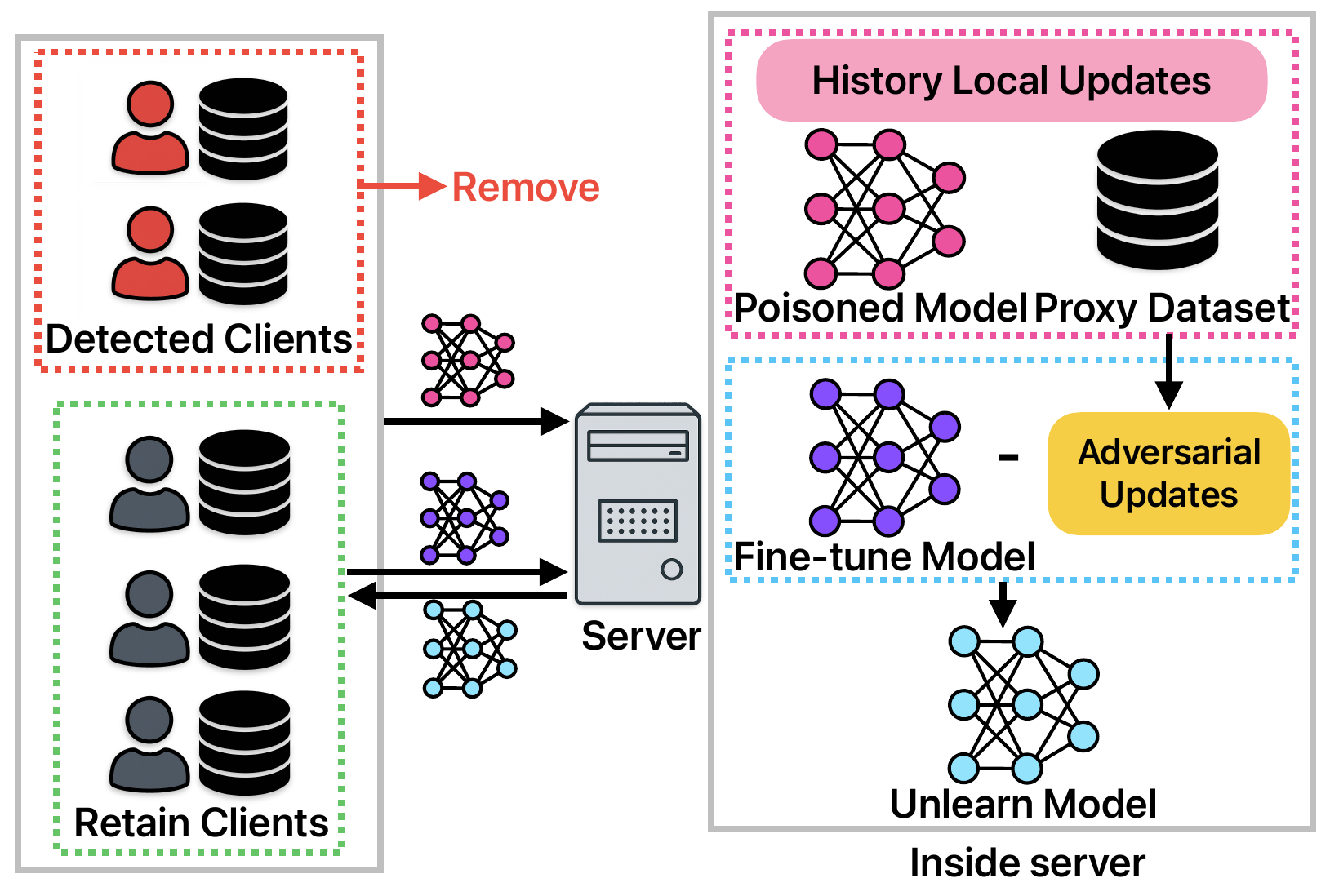} % Increase the width to make the figure larger
  \vspace{-2mm} % Adjust the space below the figure
  \caption{FAUN pipeline with adversarial updates from history updates in the first unlearning round.}
  \vspace{-5mm}
  \label{fig:FAUN}
\end{figure}

We propose an adversarial update-based federated unlearning framework that efficiently removes the influence of malicious clients from a poisoned global model. Our method FAUN exploits the server's ability to store historical client updates and to utilize a small proxy dataset for adversarial optimization. The key idea is to approximate and counteract the contribution of detected malicious clients through constructed adversarial updates, instead of retraining from scratch.  
Since malicious clients tend to introduce consistent harmful directions during training to maximize their objectives, averaging their recent updates approximates their overall influence. By further amplifying this direction with adversarial optimization on proxy data, we obtain adversarial updates that capture their worst-case effect, following a min–max principle of maximizing proxy loss while minimizing retained loss. Subtracting these adversarial updates from the poisoned model counteracts accumulated malicious influence and facilitates rapid recovery. Fig.~\ref{fig:FAUN} illustrates the first unlearning round, where adversarial updates are generated from the poisoned model; in subsequent rounds, the fine-tuned model replaces the poisoned one for generating adversarial updates.
After a limited number of adversarial elimination rounds, the server resumes fine-tuning with the retained clients, which restores model performance on benign data.

\subsection{Adversarial Update Construction}
For each malicious client $i \in \mathcal{C}_m$, the server first constructs an adversarial direction from the history of its updates before unlearning. Let $r$ index the standard FL training rounds and let $R$ denote the last training round. For unlearning, we denote the global model as $\mathbf{\hat{w}}_t$, initialized with $\mathbf{\hat{w}}_0 = \mathbf{w}_R$, where $\mathbf{w}_R$ is the poisoned model obtained at the end of training.
For each $i\in\mathcal{C}_m$, the server retrieves the last $R_m$ local updates $\{\mathbf{g}^i_{R-R_m+1},\ldots,\mathbf{g}^i_{R}\}$ and computes an intra-client averaged update $\bar{\mathbf{g}}^i=\frac{1}{R_m}\sum_{r=0}^{R_m-1}\mathbf{g}^i_{R-r}$, which captures the representative malicious behavior while smoothing round-wise fluctuations across training rounds.

To further align with the worst-case harmful effect, the server perturbs $\bar{\mathbf{g}}^i$ by an adversarial direction $\boldsymbol{\delta}^{i\ast}$ optimized on a small clean proxy dataset $\mathcal{D}_{\text{proxy}}$ via Projected Gradient Descent (PGD)~\cite{shafahi2019adversarial}:
\begin{equation}
\boldsymbol{\delta}_t^{i\ast}
=\arg\max_{\|\boldsymbol{\delta}\|_2\le\epsilon}\;
\mathcal{L}\!\big(\mathbf{\hat{w}}_t-\bar{\mathbf{g}}^i-\boldsymbol{\delta};\mathcal{D}_{\text{proxy}}\big),
\end{equation}
where $\mathbf{\hat{w}}_t$ is the global model in unlearning, and $\epsilon>0$ bounds the $\ell_2$-norm of the perturbation. The adversarial update for client $i$ is then $\mathbf{g}_t^{i\ast}=\bar{\mathbf{g}}^i+\boldsymbol{\delta}_t^{i\ast}$, and the server aggregates them into a single elimination update:
\begin{equation}
\mathbf{\bar{g}}_t^{\ast}=\frac{1}{|\mathcal{C}_m|}\sum_{i\in\mathcal{C}_m}\mathbf{g}_t^{i\ast}.
\end{equation}
At each unlearning round, the server updates the global model by combining benign aggregation with the adversarial elimination term:
\begin{equation}
\mathbf{\hat{w}}_{t+1} = \mathbf{\hat{w}}_t - \eta \cdot \mathcal{A}\!\big(\{\mathbf{\hat{g}}_t^{i}\}_{i \in \mathcal{C}_r}\big) - \lambda_t \cdot \mathbf{\bar{g}}_t^{\ast}.
\end{equation}
where $\lambda_t>0$ controls the elimination strength at round $t$. This update rule reflects a min-max principle: maximizing the proxy loss to capture worst-case malicious effects, while minimizing the retained loss to guide recovery.

\begin{figure}[t]
  \centering
  \vspace{0mm} % Adjust the space above the figure
  \includegraphics[width=0.8\linewidth]{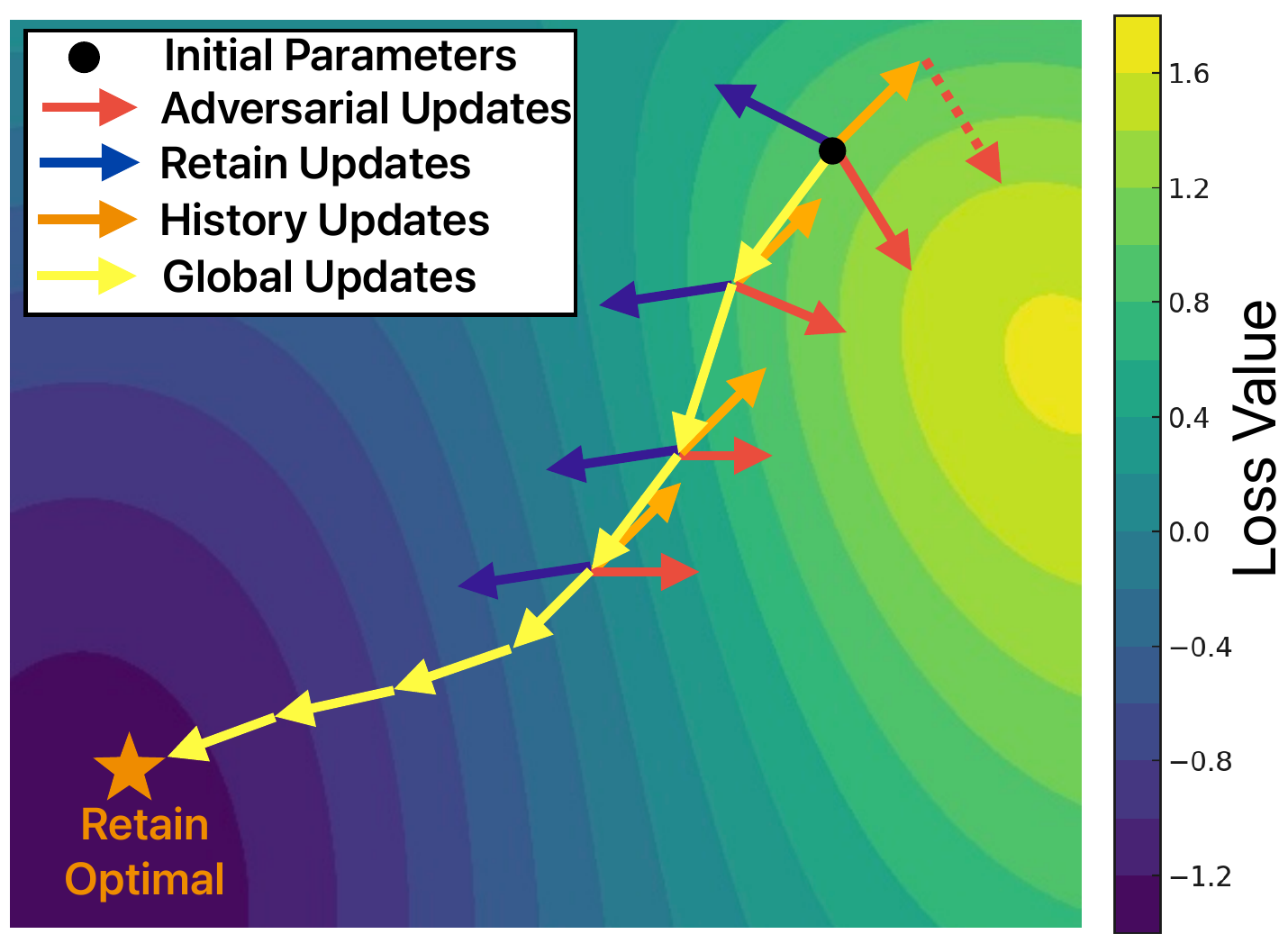} % Increase the width to make the figure larger
  \vspace{-2mm} % Adjust the space below the figure
  \caption{FAUN update dynamics on the loss landscape, where adversarial updates counteract malicious influence and retained updates guide recovery.}
  \vspace{-5mm}
  \label{fig:adv}
\end{figure}

Instead of a monotonically decaying schedule, adversarial elimination is applied only in the first $T_a$ unlearning rounds, i.e.,
\begin{equation}
\lambda_t =
\begin{cases}
1, & 0 \leq t < T_a, \\
0, & t \geq T_a,
\end{cases}
\end{equation}
so that the adversarial phase is restricted to a limited number of rounds. After $T_a$, the server resumes standard training on $\mathcal{C}_r$ to further restore performance on benign data. Fig.~\ref{fig:adv} visualizes FAUN's update trajectory on the loss landscape, where adversarial updates neutralize malicious influence and retained updates drive the model toward the benign optimum.

\section{Experiments}

\subsection{Experiment Setups}

\textbf{Datasets and model architectures:}
We conduct experiments on three datasets: MNIST~\cite{lecun1998gradient}, CIFAR-10~\cite{krizhevsky2009learning}, and AG News~\cite{ilhan2023scalefl}. MNIST and CIFAR-10 are 10-class image benchmarks, while AG News is a 4-class text classification dataset. We use a lightweight CNN~\cite{lecun2002gradient} for MNIST, ResNet-18~\cite{he2016deep} for CIFAR-10, and FastText~\cite{umer2023impact} for AG News. To emulate realistic data distributions, we consider both IID and non-IID settings. For the non-IID case, data is partitioned among clients using a Dirichlet distribution\cite{connor1969concepts} with concentration parameter 0.5 to control heterogeneity.

\textbf{Attack setup:}
Malicious clients launch poisoning attacks during training. For untargeted attacks, we adopt the Trim attack\cite{fang2020local}, which perturbs updates to evade detection while degrading the global model. For targeted attacks, we employ a backdoor strategy\cite{baruch2019little} by poisoning 10\% of local training data with predefined triggers and relabeling them to a fixed target class. Specifically, the triggers are a $3 \times 3$ white square at the bottom-right corner for MNIST, a $5 \times 5$ red “X” at the bottom-left corner for CIFAR-10, and a rare token randomly inserted into input sentences for AG News. In all cases, malicious clients performing such attacks are detected and removed from the FL system.

\textbf{Compared federated unlearning methods:}
We benchmark FAUN against three recovery methods: FedRecover\cite{cao2023fedrecover}, FedEraser\cite{liu2021federaser}, and Crab\cite{jiang2024towards}. FedRecover leverages historical gradients and models to estimate unlearning updates with a Cauchy mean value theorem. FedEraser calibrates updates using the norm of historical local updates and the direction of retrained updates. Crab extends FedEraser with round selection and client filtering to improve efficiency. We also compare with full retraining from scratch, which provides an upper bound on recovery performance at substantial cost.

\textbf{Default scenario:}
We consider MNIST with a CNN trained by SGD (learning rate 0.01, momentum 0.9) under IID data. The FL system employs FedAvg with 20 clients, where $m=6$ malicious clients are detected and $n=14$ benign clients remain for unlearning. In FAUN, we set $\epsilon=1$, PGD steps to 30 with step size 0.05, $R_m=10$, and $T_a=10$. Training lasts $R=200$ rounds with 5 local epochs per client. Since FAUN only requires including the $\lambda_t \cdot \bar{\mathbf{g}}_t^{\ast}$ term in the first $T_a$ rounds (typically $T_a<20$), the subsequent rounds serve as fine-tuning, and the total number of rounds is reduced to 50. Other unlearning methods and full retraining must start from scratch and thus require 200 rounds, consistent with the original setup. For these baselines, we adopt the default parameter settings from their respective papers.

\textbf{Evaluation metrics:}
We use multiple metrics to assess recovery performance. (1) Under Trim attack, we report model accuracy (ACC), the proportion of correctly classified test inputs, and membership inference success rate (MISR), which reflects the extent to which malicious clients' data remain memorized, a MISR close to 0.5 indicates random guessing, reflecting more complete unlearning. (2) Under backdoor attack, we measure attack success rate (ASR), the percentage of inputs with triggers misclassified into the target class, and main-task accuracy (MA), i.e., test accuracy on clean data. All reported values of ACC, MISR, ASR, and MA are expressed in percentages (\%), while we report only numerical values without the percentage sign. In addition, we record the time cost (in seconds) to evaluate the efficiency of unlearning methods.

\subsection{Experiment Results}

\textbf{FAUN performance under different datasets and model architectures:} Table~\ref{tab:dataset} summarizes the recovery results of FAUN against Trim and Backdoor attacks on three datasets and model settings. For Trim attack, we report the training-time global model accuracy (ACC), the recovered ACC using FAUN, and the ACC after retraining. As shown, FAUN effectively restores model performance across all datasets, e.g., on CIFAR-10 the ACC improves from 31.3\% to 82.2\%, and for MNIST and AG News the recovered ACC is also close to that of retraining. Similarly, for backdoor attack, Table~\ref{tab:dataset} presents attack success rate (ASR) and main-task accuracy (MA). In all cases, ASR drops below 3\% while MA remains nearly unaffected, achieving recovery comparable to retraining.Overall, FAUN achieves recovery across three datasets and two attack types that is consistently close to retraining.

\begin{table}[h]
\centering
\small % \small, \footnotesize, \scriptsize, \tiny
\setlength{\tabcolsep}{2.8pt} 
\renewcommand{\arraystretch}{0.95} 
\vspace{0cm}
\caption{FAUN performance under different datasets.}
\label{tab:dataset}
\vspace{-0.2cm}
\begin{tabular}{c|ccc|ccc}
\hline
\multirow{2}{*}{Dataset} & \multicolumn{3}{c|}{Trim Attack} & \multicolumn{3}{c}{Backdoor Attack} \\ 
                         & Train & Unlearn & Retrain        & Train & Unlearn & Retrain \\ \hline
MNIST   & 70.4  & 96.3    & 97.8           & 99.8/97.8 & 0.5/97.3   & 0.3/97.8 \\
CIFAR10 & 31.3  & 82.2    & 84.5           & 88.2/83.9 & 1.3/84.4   & 0.2/84.5 \\ 
AG News & 56.8  & 89.4    & 90.3           & 44.2/88.6 & 2.7/90.2   & 0.7/90.3 \\\hline
\end{tabular}
\vspace{0cm}
\end{table}

\textbf{Evaluation of recovery methods under poisoning attacks:}
To further analyze the advantage of FAUN, we compare it with three representative federated unlearning methods. As shown in Table~\ref{tab:dfu_methods}, FAUN achieves recovery results consistently closest to retraining under both Trim and Backdoor attacks. FAUN completes recovery in just 50 rounds (124s), while competing methods are trained for 200 rounds. This setting might seem disadvantageous to FAUN; however, even with substantially more rounds, their performance remains inferior, underscoring FAUN's efficiency advantage. Moreover, FedRecover and FedEraser must use the same number of rounds as training and all three baselines start unlearning from scratch, which naturally incurs a higher cost. In contrast, FAUN resumes from the trained global model, removes malicious influence within 20 rounds, and finishes with light fine-tuning, significantly reducing recovery time.

\begin{table}[h]
\centering
\small % \small, \footnotesize, \scriptsize
\setlength{\tabcolsep}{5.2pt} 
\renewcommand{\arraystretch}{0.95}
\vspace{0cm}
\caption{Performance of different recovery methods under Trim and Backdoor attacks.}
\label{tab:dfu_methods}
\vspace{-0.2cm}
\begin{tabular}{c|ccc|ccc}
\hline
\multirow{2}{*}{Unlearn Method} & \multicolumn{3}{c|}{Trim Attack} & \multicolumn{3}{c}{Backdoor Attack} \\ 
                                & ACC & MISR & Time & ASR & MA & Time \\ \hline
FAUN        & 96.3 & 51.9 & 124 & 0.5 & 97.3 & 124 \\
FedRecover  & 92.7 & 58.2 & 790 & 0.5 & 90.4 & 744 \\
FedEraser   & 95.3 & 69.3 & 490 & 1.1 & 91.8 & 490 \\
Crab        & 93.6 & 57.9 & 376 & 0.8 & 93.7 & 338 \\
Retrain     & 97.8 & 48.3 & 406 & 0.3 & 97.8 & 406 \\ \hline
\end{tabular}
\vspace{0cm}
\end{table}

\begin{figure}[t]
  \centering
  % left
  \begin{minipage}[b]{0.48\linewidth}
    \centering
    \includegraphics[width=\linewidth]{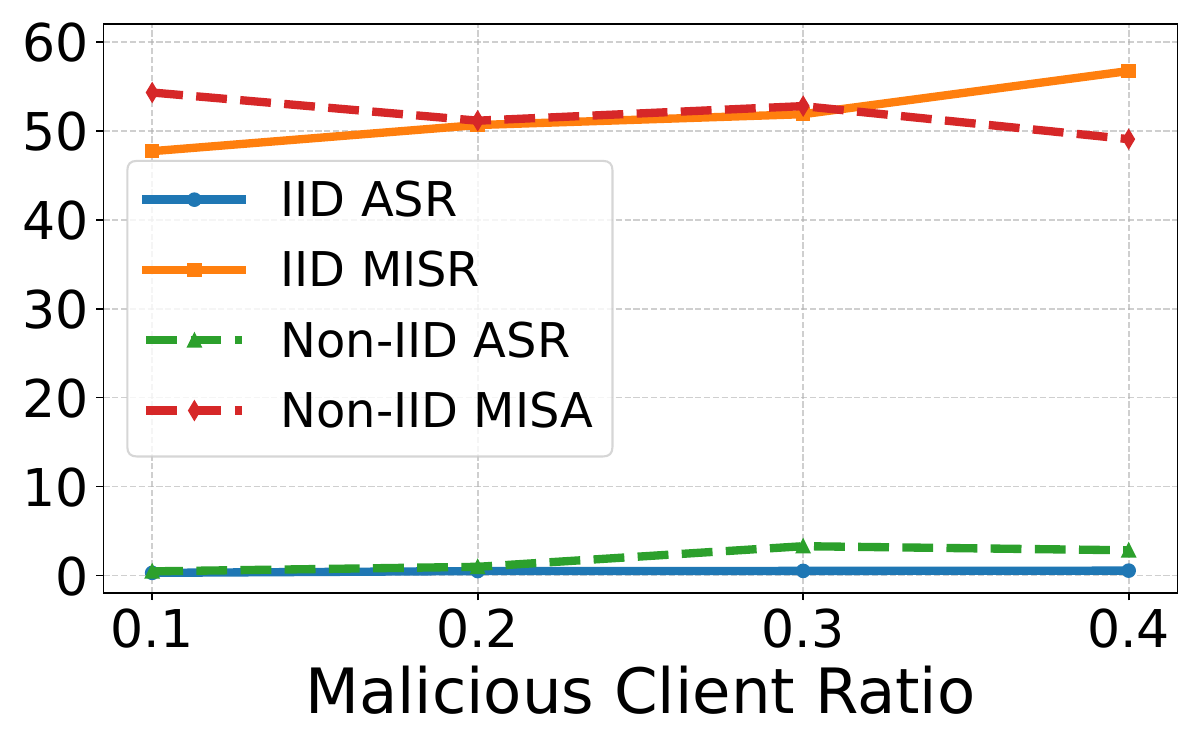}
    \vspace{-0.7cm}
    \caption{FAUN performance under varying malicious client ratios in data IID and Non-IID settings.}
    \label{fig:noniid}
  \end{minipage}
  \hfill
  % right
  \begin{minipage}[b]{0.48\linewidth}
    \centering
    \includegraphics[width=\linewidth]{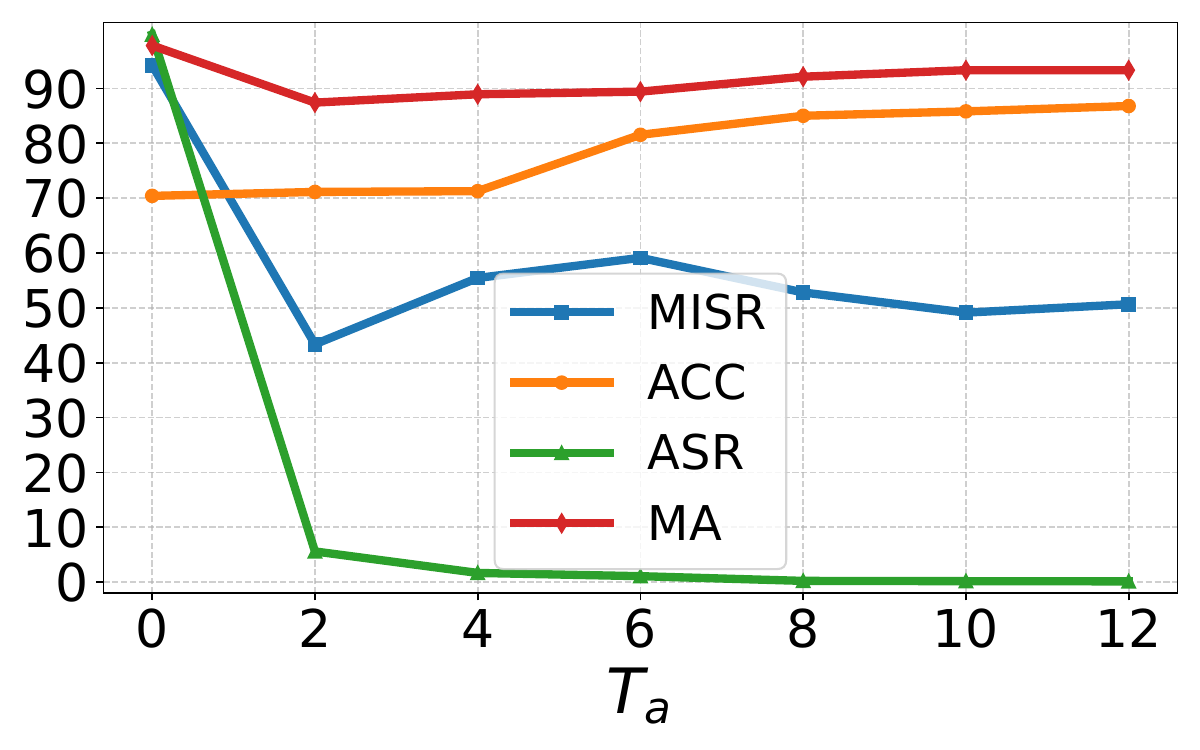}
    \vspace{-0.7cm}
    \caption{Impact of adversarial elimination rounds $T_a$ on FAUN performance in terms of MISR and ASR.}
    \label{fig:Ta}
  \end{minipage}
\vspace{-0.5cm}
\end{figure}

\textbf{Evaluation of different malicious ratio and noniid data distribution:}
Fig.~\ref{fig:noniid} illustrates FAUN's recovery performance across different ratios of malicious clients. Since previous results have shown that FAUN preserves model accuracy comparable to retraining, we report MISR to evaluate recovery under Trim attacks and ASR to assess backdoor removal. FAUN consistently achieves MISR around 50\% and ASR below 3\% across all malicious ratios. Similar trends are observed in both IID and non-IID settings, demonstrating that FAUN effectively mitigates both untargeted and targeted attacks regardless of data distribution.

\textbf{Impact of adversarial elimination rounds $T_a$:}
Our method removes poisoning information through adversarial updates, while model performance is mainly restored by fine-tuning. We evaluate the effect of adversarial updates within $T_a$ rounds on eliminating poisoned information. As shown in Fig.~\ref{fig:Ta}, only two rounds already reduce MISR to around 50\% and ASR below 10\%. With larger $T_a$, MISR converges closer to 50\% and ASR drops below 1\% once $T_a \geq 8$. This demonstrates that adversarial updates effectively mitigate both Trim and Backdoor attacks, with elimination largely completed by $T_a \geq 10$. Excessive rounds are unnecessary as they incur extra overhead from computing adversarial updates. Meanwhile, ACC and MA increase slowly with $T_a$ and mainly rely on fine-tuning for recovery; however, as they already reach high values, only a few fine-tuning rounds are required to fully restore performance.

\section{Conclusion}

We introduced Federated Adversarial Unlearning (FAUN), a method for recovering federated learning models after poisoning attacks. FAUN constructs adversarial updates from a short window of malicious clients' updates and a small proxy dataset to capture and eliminate worst-case malicious directions. Limiting adversarial elimination to a few rounds and then resuming training with benign clients enables both fast removal of malicious effects and stable recovery. Experiments on image and text benchmarks under untargeted and targeted poisoning show that FAUN achieves recovery comparable to retraining while greatly reducing computation, rounds, and storage. These results demonstrate adversarial-update-based unlearning as an effective and efficient paradigm for mitigating poisoning in federated learning.
\noindent{\bf Acknowledgement:} The work was supported in part by NSF Grant 2319781 at University of South Florida.

\bibliographystyle{IEEEbib}
\bibliography{references}

\end{document}